# Can OpenAI o1 Reason Well in Ophthalmology? A 6,990-Question Head-to-Head Evaluation Study

Sahana Srinivasan[1,2]*, Xuguang Ai[3]*, Minjie Zou[1]*, Ke Zou[1], Hyunjae Kim[3], Thaddaeus Wai Soon Lo[2], Krithi Pushpanathan[1], Yiming Kong[3], Anran Li[3], Maxwell Singer[4], Kai Jin[5], Fares Antaki[6,7], David Ziyou Chen[1,8], Dianbo Liu[1], Ron A. Adelman[4], Qingyu Chen[3]†, Yih Chung Tham[1,2,9]†

1. Centre for Innovation and Precision Eye Health, Department of Ophthalmology, Yong Loo Lin School of Medicine, National University of Singapore, Singapore.
2. Singapore Eye Research Institute, Singapore National Eye Centre, Singapore
3. Department of Biomedical Informatics and Data Science, Yale School of Medicine, Yale University, New Haven, USA
4. Department of Ophthalmology and Visual Science, Yale School of Medicine, Yale University, New Haven, USA
5. Eye Center, The Second Affiliated Hospital, School of Medicine, Zhejiang University, Hangzhou, Zhejiang, China.
6. Cole Eye Institute, Cleveland Clinic, Cleveland, OH, USA
7. The CHUM School of Artificial Intelligence in Healthcare, Montreal, QC, Canada
8. Department of Ophthalmology, National University Hospital, Singapore.
9. Eye Academic Clinical Program (Eye ACP), Duke NUS Medical School, Singapore.

*contributed equally as first authors

†contributed equally as last authors

**Key Points**

Question:

What is the performance and reasoning ability of OpenAI o1 compared to other large language models in addressing ophthalmology-specific questions?

Findings:

This study evaluated OpenAI o1 and five LLMs using 6,990 ophthalmological questions from MedMCQA. O1 achieved the highest accuracy (0.88) and macro-F1 score but ranked third in reasoning capabilities based on text-generation metrics. Across subtopics, o1 ranked first in "Lens" and "Glaucoma" but second to GPT-4o in "Corneal and External Diseases," "Vitreous and Retina," and "Oculoplastic and Orbital Diseases". Subgroup analyses showed o1 performed better on queries with longer ground truth explanations.

Meaning:

O1's reasoning enhancements may not fully extend to ophthalmology, underscoring the need for domain-specific refinements to optimize performance in specialized fields like ophthalmology.

Abstract:

Importance: OpenAI's recent large language model (LLM), o1, has been designed with enhanced reasoning capabilities. However, this advancement remains untested in specialised medical fields like ophthalmology. Findings from this study provide valuable insights into o1's performance in ophthalmology, addressing whether a domain-specific LLM is still needed.

Objective: To determine the performance and reasoning ability of OpenAI o1 compared to other commonly used LLMs in addressing ophthalmology-specific questions.

Design and Setting: This evaluation included Open AI's o1, GPT-4o, GPT-4, GPT-3.5, along with Llama 3, and Gemini 1.5 Pro. A total of 6,990 questions from the ophthalmology subset of the open-source MedMCQA examination dataset was input into each LLM in a standardised format to assess medical knowledge accuracy and reasoning. Subgroup analyses were performed based on ophthalmic subtopics and ground truth explanation lengths.

Main Outcomes and Measures: The models were evaluated on performance (using accuracy and macro-F1 score) and reasoning abilities (measured using text-generation metrics: ROUGE-L, BERTScore, BARTScore, AlignScore, METEOR, and a weighted normalised aggregate score).

Results: Among the six models, o1 achieved the highest accuracy (0.88 ± 0.33) and macro-F1 score (0.70) (all P<0.001). However, in reasoning evaluation, based on the weighted aggregate score, o1 ranked third behind GPT-4o and GPT-4.  O1 performed poorer than GPT4o and GPT-4 in ROUGE-L (0.12 ± 0.07), BERTScore (0.66 ± 0.06), and AlignScore (0.16 ± 0.13) (all P<0.001). Nevertheless, O1 excelled in BARTScore (-4.79 ± 1.09) and METEOR (0.22 ± 0.10). Across subtopics, o1 ranked first in "Lens" and "Glaucoma" but was predominantly second to GPT-4o in "Corneal and External Diseases," "Vitreous and Retina," and "Oculoplastic and Orbital Diseases." Additionally, o1's reasoning showed improved performance on MCQ items with longer ground truth explanations (≥100 words).

Conclusions and Relevance:  OpenAI o1 demonstrated superior accuracy in answering ophthalmology-related MCQs. However, its reasoning abilities lagged behind those of GPT-4o and GPT-4. These findings suggest that o1's general reasoning enhancements may not extend equivalently to ophthalmology, underscoring the ongoing need for domain-specific refinements to optimise LLM performance in specialised medical fields like ophthalmology.

**Introduction**

The emergence of large language models (LLMs) represents a transformative advancement in natural language processing (NLP). LLMs generate "human-like" responses by prediction of the natural language token based on their extensive training.[1] Since the introduction of OpenAI's GPT-3.5 in late 2022,[1–6] LLMs have made significant advancements the last two years. In healthcare, LLMs have shown promise to revolutionise or optimise current clinical practice and medical education delivery.[3,5–11] For example, LLMs have shown potential in simplifying and summarising medical reports in radiology and intensive care, triaging in emergency medicine and answering patient health queries. [12–15]

In the field of ophthalmology, LLMs have demonstrated potential utility in improving healthcare delivery by optimising disease diagnosis, patient education and clinical workflow.[16–21] However, a critical limitation of existing LLMs lies in their susceptibility to generating inaccurate outputs, and their inconsistent performance in medical-specific tasks.[1,5,6,22–25] For this reason, LLMs are constantly being improved to generate outputs that are more accurate and with more up-to-date knowledge.

The rapid development of LLMs has culminated in the release of OpenAI o1 in September 2024. This recent iteration is designed with enhanced reasoning capabilities, aimed at tackling complex tasks with greater accuracy.[26] This iteration retains the hallmark features of earlier LLMs such as "human-like" conversation while introducing advancements in reasoning and problem-solving processes.[26,27] With a longer time for reasoning and response generation, it is claimed that o1 can now "think before it responds".[27] This enhancement positions o1 as a model anticipated to outperform its predecessors, potentially enabling it to handle more complex tasks.[26,28] Nevertheless, while recent studies demonstrated o1's improved performance in answering general medicine-related questions,[28–31] its reasoning and problem-solving performance in the specialised field, in particular ophthalmology remains unexplored.

To address this gap, this study aimed to evaluate and compare the performance of OpenAI o1 with five other commonly used LLMs - GPT-4o, GPT-4, GPT-3.5, Llama-3-8b and Gemini 1.5 Pro, in addressing ophthalmology-related questions. An ophthalmological exam benchmark dataset was first curated by extracting ophthalmological questions from the Medical Multiple-Choice Question Answering (MedMCQA) dataset. The models were then comprehensively assessed for both the accuracy of their responses and the quality of their reasoning. These findings may offer valuable insights into the capabilities and limitations of the o1 model in ophthalmology and contribute to answering the critical question of whether a domain-customised, ophthalmology-specific LLM is still necessary.

**Methods**

This study was a comparative evaluation of six LLMs, o1, GPT-4o, GPT-4, GPT-3.5, Llama-3-8b, and Gemini 1.5 Pro, assessing their performance on a set of multiple-choice questions (MCQs) using both accuracy- and reasoning-focused metrics.

Models evaluated

The models included in the analysis were OpenAI's o1 (o1-preview-2024-09-12), GPT-4o (gpt-4o-2024-05-13), GPT-4 (gpt-4-0613), GPT-3.5 (gpt-35-turbo-0613, as well as Llama-3-8b (Meta-Llama-3-8B-Instruct) and Gemini 1.5 Pro (gemini-1.5-pro-001). All models were accessed via their respective APIs.

Dataset

This study utilized the ophthalmology subset of the MedMCQA dataset, which comprised of 6,990 multiple-choice questions (MCQs).[32] Each MCQ item consisted of a correct answer and an accompanying explanation, which served as the ground truth reasoning for comparison with the outputs generated by the various LLMs in this study.

Prompt Engineering and hyperparameters

Since prompt phrasing can influence model performance, we standardised the input format, consistent with the styles adopted in previous studies.[33,34] As illustrated in **Figure 1**, the 6,990 questions were reformatted into prompts, before being presented to each model with a zero-shot approach. Each prompt instructed the model to respond as an "assistant specializing in ophthalmology", select the most appropriate answer option, and provide an explanation for its choice (**Figure 1**). We conducted greedy decoding with a temperature set to 0 across all models to minimise response variability.[35]

Evaluation Metrics

We evaluated outputs of o1 and the five other benchmark LLMs using a variety of metrics to assess accuracy and reasoning. Accuracy was evaluated using traditional accuracy which was defined as the proportion of correct predictions made out of the total predictions mad, while macro-F1 score was used to provide the unweighted harmonic mean of precision and recall. Reasoning abilities were evaluated using five text-generation metrics: Recall-Oriented Understudy for Gisting Evaluation (ROUGE-L), BERTScore, BARTScore, AlignScore and the Metric for Evaluation of Translation with Explicit Ordering (METEOR).[36–40] These metrics evaluate the quality of the model's generated reasoning by comparing it to the ground truth (reference) reasoning in MedMCQA, assessing aspects such as similarity and overlap in key phrases and words (See **Supplementary Table 1** for definitions). To summarise reasoning performance, a weighted normalised aggregate score was calculated. For each text-generation metric, scores were first normalised on a scale from 0 to 1, with 0 representing the poorest performance and 1 being the best. The normalized scores across the five metrics were then averaged to derive a final weighted score for each LLM.

Subgroup Evaluations

We conducted three subgroup or sensitivity analyses to further explore how different factors may influence o1's reasoning capabilities. The ophthalmology subset of the MedMCQA dataset, with its vast 6,990 items, provides a comprehensive resource for evaluating LLM performance. However, it is unavoidable that some ground truth explanations may be suboptimal. For instance, we observed a small subset of items where explanations merely listed the answer options, restated the correct answer verbatim, or consisted of disjointed keywords rather than coherent reasoning (**Supplementary Table 2**). Such suboptimal reasoning could confound performance on text-generation metrics, which rely on comparing model outputs to reference explanations. To address this, we conducted a sensitivity analysis on 100 questions with complete and detailed explanations for both correct and incorrect options.

Second, we additionally performed subgroup analyses stratified by ground truth explanations' length: i) those with ≥50 words, ii) ≥100 words.

Thirdly, we explored whether o1's performance varied across various ophthalmic sub-topics. The original ophthalmology subset of MedMCQA contained 123 overlapping subtopics. For better clarity, we regrouped the 123 sub-topics into distinct topic groups (**Supplementary Table 3**). To ensure robust evaluation, we then focused on five ophthalmic topic groups with the largest sample sizes: "Corneal and external diseases" (n=534), "Lens" (n=336), "Vitreous and retina" (n=319), "Oculoplastic and orbital diseases" (n=256), and "Glaucoma" (n=213).

Subset with Human Expert Qualitative Review

Text-generation metrics alone may not entirely reflect the reasoning abilities of these models.[25,41] To address this, a qualitative review was conducted by two human evaluators (SS, MJW) on a randomly selected subset of 40 items. Specifically, the evaluators assessed responses generated by OpenAI o1 and its predecessor, GPT-4o, to identify similarities or differences in reasoning quality and coherence.

Statistical Analysis

All statistical analyses were conducted using Python (Python version 3.9.0, Python Software Foundation), with statistical significance set at p<0.05. To compare accuracy, the paired t-test was used to assess any statistical significance between o1 and the other models' performances. The z-test was used for macro-F1 as it provides a single aggregate performance score. Although the distribution of the dataset is non-parametric, its large sample size (n=6,990) allows for the difference in means to be approximated as a normal distribution, justifying the use of the paired t-test and z-test. Additionally, the Bonferroni correction was applied to control the family-wise error rate in multiple test comparisons.

For the text-generation metrics, the two-tailed Wilcoxon rank-sum test was used to assess statistical significance between o1 and the other models' performances at the item level. This test was chosen due to the non-parametric nature of the text-generation metrics' output value distributions. The Bonferroni correction was applied here as well to adjust for multiple comparisons.

## Results

### Performance on Accuracy and Macro-F1 Scores

**Figure 2a** and **Table 1** highlight OpenAI o1's performance in accuracy and macro-F1 scores. Overall, o1 demonstrated the highest accuracy (0.88 ± 0.33) among all models, significantly outperforming the others (all $p < 0.001$). It was followed by Llama-3-8b (0.83± 0.38, GPT-4o (0.82± 0.39), GPT-4 (0.75± 0.43), Gemini 1.5 Pro (0.71± 0.45), and GPT-3.5 (0.58± 0.49). Similarly, o1 achieved the highest macro-F1 score (0.70), followed by Llama-3-8b (0.66), GPT-4o (0.65), GPT-4 (0.60), Gemini 1.5 Pro (0.57), and GPT-3.5 (0.46) (all $p < 0.001$).

### Performance on Text-Generation Metrics

In reasoning evaluations, based on the weighted normalised score - o1 ranked third overall, behind GPT-4o (first) and GPT-4 (second) (**Figure 2b** and **Table 2**). Based on ROUGE-L score, o1 (0.12±0.07) ranked fourth, behind GPT-4 (0.15±0.08), GPT-4o (0.14±0.07), and GPT-3.5 (0.13±0.07). Similarly, for BERTScore, o1 (0.66±0.06) ranked third, behind GPT-4o (0.67±0.05) and GPT-4 (0.67±0.05) (all p<0.001). A similar trend was observed for AlignScore (o1: 0.16±0.13, GPT-4o: 0.18±0.14, GPT-4: 0.19±0.15; p<0.001). However, o1 excelled in BARTScore, achieving the best performance with a score of -4.79±1.09, outperforming all other models (all p<0.001). For METEOR, o1's performance (0.22±0.10) was joint-best with GPT-4o (p=0.06) and significantly surpassed the other models (all p<0.001). In an additional sensitivity analysis where we only included MCQ items which were correctly answered by all LLMs, we consistently observed that o1 ranked first for BARTScore and METEOR, while ranking behind GPT-4 (first) and GPT-4o (second) for ROUGE-L, BERTScore and AlignScore (all p≤0.01, except for comparison in BERTScore). Similarly, o1 performed comparably to Llama-3-8b which ranked third for ROUGE-L and BERTScore **(Supplementary Table 4).** In a sensitivity analysis which only evaluated MCQ items which had ground truth explanations for both correct and incorrect answers, similar trends were observed as well (**Supplementary Table 5**).

**Subgroup Analyses by Ground Truth Explanation Length**

Sub-analyses were conducted on questions with ground truth explanations of ≥50 words and ≥100 words (**Table 3**). Scores across all metrics generally improved as the ground truth explanation length increased, Notably, o1 outperformed the other models when the ground truth explanation length exceeded 100 words, achieving the highest scores in BERTScore, BARTScore, and METEOR (p ≤0.001 for all comparisons, except with GPT-4o in BERTScore). Similarly, o1 outperformed GPT-4 and Gemini 1.5 Pro in ROUGE-L and tied first with the rest (P<0.001) However, for AlignScore, even when the ground truth explanation length was ≥100 words, o1 (0.30±0.12) still ranked third behind GPT-4o (0.33±0.14) and GPT-4 (0.33±0.15).

**Performance Across Ophthalmic Subtopics**

When evaluating o1's performance across the five ophthalmic topic groups, the model consistently achieved the highest accuracy, largely outperforming all other models (**Supplementary Table 6**).

O1's performance varied across the text-generation metrics in different topic groups, mirroring trends seen in the entire dataset (**Supplementary Table 7**). Overall, using the weighted normalised scores, o1 ranked first in "Lens" (o1: 0.75, others: 0.05-0.65) and "Glaucoma" (o1: 0.76, others: 0.13-0.72). However, overall, o1 ranked second to GPT-4o in "Corneal and external diseases" (o1: 0.75, GPT-4o: 0.76), "Vitreous and retina" (o1: 0.73, GPT-4o: 0.80), and "Oculoplastic and orbital diseases" (o1: 0.74, GPT-4o: 0.76). Across the topic groups, o1 consistently ranked first in terms of BARTScore and METEOR, but still ranked second or third behind GPT4o and GPT4 for the ROUGE-L, BERTScore and AlignScore metrics.

**Human Expert Review of LLM-generated Responses**

This review on a randomly selected subset of MCQ items revealed distinct response patterns between o1 and GPT-4o. We observed that o1 produced more structured and detailed responses (**Supplementary Table 8a**), maintaining a methodical format even for incorrect outputs (**Supplementary Table 8b**). Additionally, o1 also incorporated more clinical terminology, aligning with its higher accuracy scores. In contrast, GPT-4o tended to generate concise, summary-like responses (**Supplementary Tables 8a and 8b**). However, qualitative evaluation revealed instances where o1 occasionally confused similar yet distinct clinical concepts or methodologies, leading to inaccurate outputs. For instance, o1 seemed to have misinterpreted phenol red thread test (PRTT) to Schirmer-2 test, leading to choosing the incorrect option when asked about characteristics of PRTT (**Supplementary Tables 8b and 8d**).

**Discussion**

Using a large dataset of 6,990 ophthalmology-related MCQ items, our study pioneers the evaluation of OpenAI's latest o1 model's performance and reasoning capabilities, benchmarking it against five other commonly used LLMs. While recent studies primarily examined o1 in general medical discipline and its accuracy performance[28,29], our study distinguishes itself through a more comprehensive and multifaceted evaluation approach which spanned across performance assessment (accuracy and macro-F1 scores), reasoning capabilities (using standard text-generation metrics), and qualitative evaluation via human expert reviews. Altogether, these methodologies provided a multidimensional assessment of o1's domain-specific knowledge and inferential reasoning processes in ophthalmology. We found that o1 demonstrated superior accuracy but largely ranked behind GPT-4o and GPT-4 in terms of the text-generation metrics. Contrary to expectations that o1's enhanced reasoning capabilities would extend to ophthalmology, our findings suggest that o1's reasoning capability may not translate equivalently to ophthalmology. This underscores the ongoing need for customised ophthalmology-specific LLMs to optimize performance in this specialised field.

In contrast to previous study, o1's ranking as third behind GPT-4o and GPT-4 in text-generation metrics in this ophthalmological evaluation differs from earlier reported findings. For instance, in an evaluation involving general medical contexts, Xie et al., demonstrated o1's superior performance in ROUGE-1 compared to other general LLMs.[29] This discrepancy could be due to the highly specialised nature of ophthalmology, where general reasoning improvements may not translate directly. Moreover, o1 tended to generate longer responses compared to other models and the ground truth explanations, likely as a result of its chain-of-thought reasoning approach.[26,27] On this note, text-generation metrics, which primarily evaluate lexical and semantic similarity between model outputs and ground truth explanations, are generally sensitive to length mismatches. Thus, overly verbose outputs may reduce similarity scores in these text-generation metrics, partly explaining o1's inconsistent performance across these text-generation metrics. However, in subgroup analyses based on ground truth explanation length, o1 showed improved performance on queries with ground truth reasonings exceeding 100 words. Taken together, these observations warrant the need for further investigations into the reasoning processes.

In our qualitative review of the randomly selected subset of MCQ items, o1 was observed to produce more detailed and methodical responses compared to GPT-4o, particularly when o1 selected the correct answer. For instance, in **Supplementary Table 8a**, o1 provided a more comprehensive breakdown of the visual cascade compared to GPT-4o, demonstrating a comparatively deeper grasp of ophthalmological knowledge. Similarly, in **Supplementary Table 8c**, o1 accurately identified key details in a step-by-step manner, leading to the correct answer. These examples highlight how o1's chain-of-thought reasoning can enhance its ability to address complex medical scenarios.

However, o1's detail-oriented approach is not without challenges. In some instances, misplaced assumptions led to errors (as illustrated in **Supplementary Table 8b)**, where o1 confused two similar tests for dry eyes, resulting in an incorrect output. Despite these limitations, o1's detailed approach provides insights into the model's reasoning process and highlights opportunities for targeted refinement in the future, by incorporating ophthalmology-specific datasets for fine-tuning. In contrast, GPT-4o produces more succinct, general responses, prioritising clarity and coherence, often avoiding over-interpretation (**Supplementary Table 8b**). While this succinctness mitigates risks of over-analysis, it may omit important medical nuances necessary for accurate judgment in complex cases (**Supplementary Table 8c, Supplementary Table 8d**).

The detailed clinical language and step-by-step reasoning in o1's responses suggest potential for tasks requiring depth and technical accuracy, such as medical staff training, patient education, or for providing second opinions for patients (**Supplementary Table 8a**). However, its chain-of-thought approach also presents challenges. Notably, this approach was associated with slower response times, increased token usage and cost compared to other models.[29] Hence, this detailed qualitative review also highlight a trade-off between the depth of o1's responses and their practical utility in different clinical scenarios. Furthermore, in this study, overall, we did not observe notable reasoning improvements of o1 over its previous iteration GPT-4o. This raises questions about o1's efficiency, practical applicability, and cost-effectiveness, particularly in time-sensitive ophthalmological practice, where speed and precision are critical.

Our study has several key strengths. Notably, it pioneers a comprehensive and multidimensional evaluation of o1 in ophthalmology. Secondly, to our knowledge, we utilised the largest ophthalmological MCQ dataset to date, comprising of 6,990 questions, providing a robust foundation for assessing model accuracy and reasoning. Third, we conducted a thorough head-to-head comparison of o1 with five other general LLMs, offering a detailed performance benchmark. Furthermore, our assessment extended beyond accuracy, incorporating five distinct text-generation metrics to provide a multi-faceted evaluations of o1's strengths and limitations. Lastly, our detailed qualitative review of 40 randomly selected outputs offered qualitative insights into the differences in reasoning patterns and outputs between o1 and GPT-4o.

Nevertheless, this study has a few limitations. Firstly, the five text-generation metrics used in this study are general and not specific to medical context. This warrants the need for medical-specific text generation metric for future study. Secondly, the use of MedMCQA's examination questions forms only the first step towards understanding the applicability of LLMs in ophthalmological clinical practice. Moving forward, there is a need for benchmarks that can be used for the systematic evaluation of their clinical value[42]. Future studies should aim to create standardised and diverse ophthalmology-specific validation datasets by curating diverse tasks, further including patient queries, and clinical management tasks[25,42].

**Conclusion**

O1 demonstrated superior accuracy in answering ophthalmology-related MCQs but showed inconsistencies in reasoning capabilities, largely ranking behind GPT-4o and GPT-4 in text-generation metrics. Contrary to expectations that o1's enhanced reasoning capabilities would extend to ophthalmology, its performance highlighted limitations in addressing domain-specific challenges. These findings suggest that a customised ophthalmology-specific LLM may still be necessary and underscore the importance of domain-specific evaluations.

**Tables**

Table 1: Comparative Performance of OpenAI o1 and Five Other Large Language Models in Accuracy and Macro-F1 Scores

| Metrics | Models | | | | | |
|---|---|---|---|---|---|---|
| | **OpenAI O1** | GPT-4o | GPT-4 | GPT-3.5 | Llama-3-8b | Gemini 1.5 Pro |
| **Accuracy** | **0.88±0.33** | 0.82±0.39** | 0.75±0.43** | 0.58±0.49** | 0.83±0.38** | 0.71±0.45** |
| **Macro-F1** | **0.70** | 0.65** | 0.60** | 0.46** | 0.66** | 0.57** |

**Denotes p-value<0.001 (after Bonferroni correction) for comparisons against OpenAI o1's performance

Table 2: Comparative Performance of OpenAI o1 and Five Other Large Language Models based on Text Generation Metrics

| Metrics | Models | | | | | |
|---|---|---|---|---|---|---|
| | OpenAI O1 | GPT-4o | GPT-4 | GPT-3.5 | Llama-3-8b | Gemini 1.5 Pro |
| ROUGE-L | 0.12±0.07 | 0.14±0.07** | **0.15±0.08**** | 0.13±0.07** | 0.12±0.07* | 0.11±0.07** |
| BERTScore | 0.66±0.06 | **0.67±0.05**** | **0.67±0.05**** | 0.66±0.05** | 0.66±0.05 | 0.64±0.06** |
| BARTScore | **-4.79±1.09** | -4.87±1.10** | -4.95±1.11** | -4.95±1.09** | -4.88±1.07** | -4.94±1.07** |
| AlignScore | 0.16±0.13 | 0.18±0.14** | **0.19±0.15**** | 0.13±0.12** | 0.15±0.12** | 0.10±0.10** |
| METEOR | **0.22±0.10** | **0.22±0.10** | 0.21±0.10** | 0.21±0.09** | 0.21±0.09** | 0.20±0.09** |
| Weighted Normalised Scores⸸ | 0.72 | **0.83** | 0.74 | 0.38 | 0.47 | 0.02 |

*Denotes p-value<0.05 (after Bonferroni correction) and **denotes p-value<0.001 (after Bonferroni correction) for comparisons against OpenAI o1's performance.

⸸Model performance in each text generation metric was normalized on a scale from 0 to 1, with 0 being assigned to the poorest score and 1 to the greatest, with the rest normalized proportionally to that scale. The weighted normalized scores were calculated by taking the mean of all the scores across the five text generation metrics.

Table 3: Subgroup Analysis of OpenAI o1 and Five Other Models Stratified by Ground Truth Explanation Length

| Sub Group | Models | | | | | |
|---|---|---|---|---|---|---|
| | **OpenAI O1** | **GPT-4o** | **GPT-4** | **GPT-3.5** | **Llama-3-8b** | **Gemini 1.5 Pro** |
| **ROUGE-L** | | | | | | |
| All questions (n=6,990) | 0.12±0.070 | 0.14±0.07** | **0.15±0.08**** | 0.13±0.07** | 0.12±0.07* | 0.11±0.07** |
| Questions with Ground Truth Explanations ≥ 50 Words (n=3,002) | 0.17±0.05 | **0.18±0.05**** | **0.18±0.06**** | **0.18±0.05**** | **0.18±0.05*** | 0.16±0.05** |
| Questions with Ground Truth Explanations ≥ 100 Words (n=1,319) | **0.18±0.04** | **0.18±0.05** | 0.17±0.05** | **0.18±0.05** | **0.18±0.04** | 0.16±0.04** |
| **BERTScore** | | | | | | |
| All questions (n=6,990) | 0.66±0.06 | **0.67±0.05**** | **0.67±0.05**** | 0.66±0.05** | 0.66±0.05 | 0.64±0.06** |
| Questions with Ground Truth Explanations ≥ 50 Words (n=3,002) | 0.69±0.03 | **0.70±0.03**** | 0.69±0.04 | 0.69±0.03** | 0.69±0.03** | 0.67±0.03** |
| Questions with Ground Truth Explanations ≥ 100 Words (n=1,319) | **0.69±0.03** | **0.69±0.03** | 0.69±0.03** | 0.69±0.03** | 0.69±0.03** | 0.68±0.03** |
| **BARTScore** | | | | | | |
| All questions (n=6,990) | **-4.79±1.09** | -4.87±1.10** | -4.95±1.11** | -4.95±1.09** | -4.88±1.07** | -4.94±1.07** |
| Questions with Ground Truth Explanations ≥ 50 Words (n=3,002) | **-4.67±0.69** | -4.79±0.67** | -4.88±0.67** | -4.82±0.67** | -4.82±0.67** | -4.8±0.67** |
| Questions with Ground Truth Explanations ≥ 100 Words (n=1,319) | **-4.70±0.63** | -4.82±0.62** | -4.90±0.62** | -4.83±0.62** | -4.83±0.62** | -4.8±0.62** |
| **AlignScore** | | | | | | |
| All questions (n=6,990) | 0.16±0.13 | 0.18±0.14** | **0.19±0.15**** | 0.13±0.12** | 0.15±0.12** | 0.10±0.10** |
| Questions with Ground Truth Explanations ≥ 50 Words (n=3,002) | **0.25±0.12** | 0.22±0.08** | 0.20±0.08** | 0.23±0.07** | 0.23±0.07** | 0.24±0.07** |
| Questions with Ground Truth Explanations ≥ 100 Words (n=1,319) | 0.30±0.12 | **0.33±0.14**** | **0.33±0.15**** | 0.24±0.13** | 0.27±0.12** | 0.21±0.11** |
| **METEOR** | | | | | | |
| All questions (n=6,990) | **0.22±0.10** | **0.22±0.10** | 0.21±0.10** | 0.21±0.09** | 0.21±0.09** | 0.20±0.09** |
| Questions with Ground Truth Explanations ≥ 50 Words (n=3,002) | **0.25±0.07** | 0.22±0.08** | 0.20±0.08** | 0.23±0.07** | 0.23±0.07** | 0.24±0.07** |
| Questions with Ground Truth Explanations ≥ 100 Words (n=1,319) | **0.23±0.07** | 0.18±0.06** | 0.16±0.06** | 0.20±0.07** | 0.21±0.06** | 0.22±0.06* |

*Denotes p-value<0.05 (after Bonferroni correction) and **denotes p-value<0.001 (after Bonferroni correction) for comparisons against OpenAI o1's performance

# Figures

## Figure 1: Standardized Prompt Format Used for Each MCQ Item

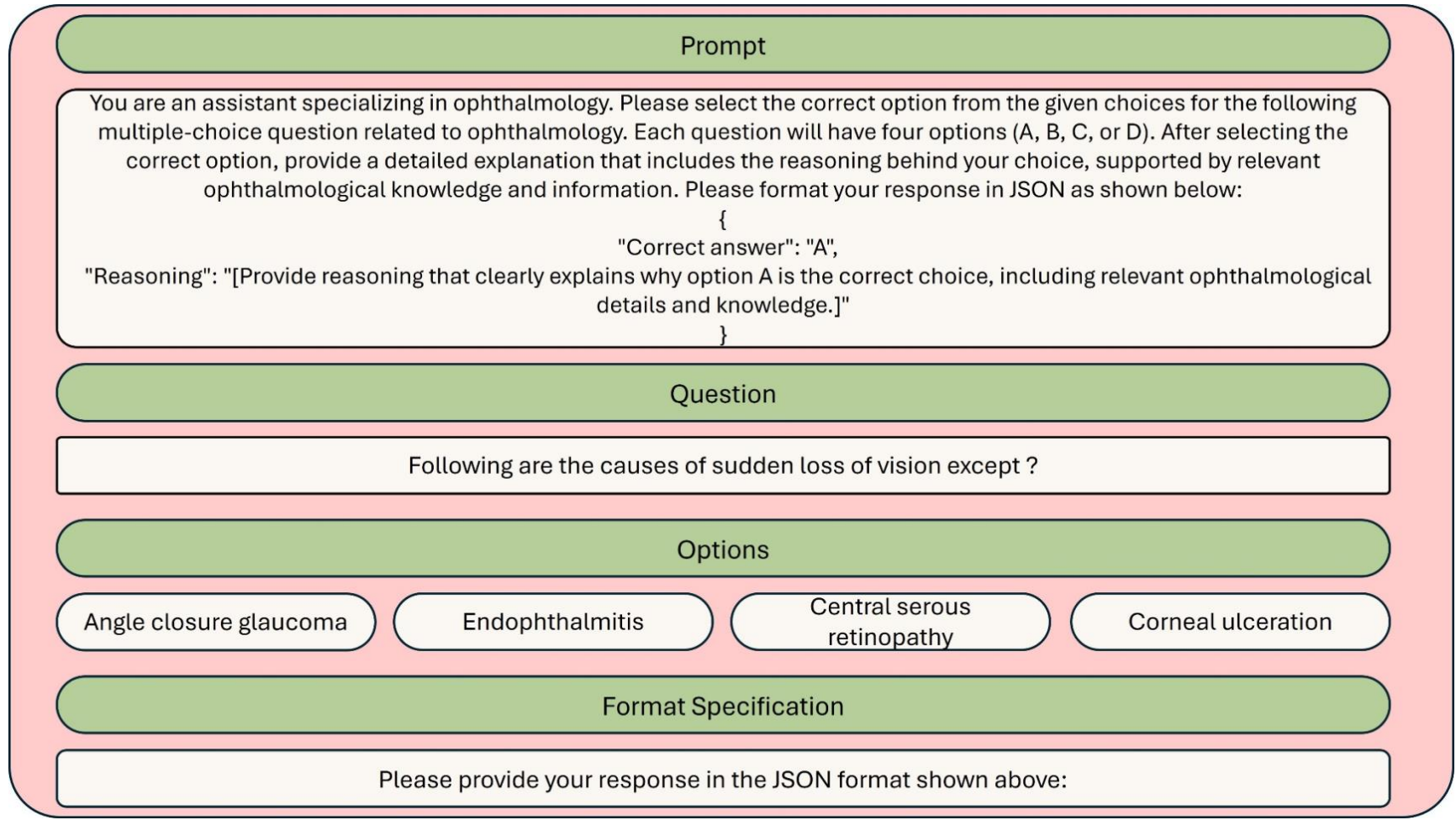

Figure 2: Comparison of OpenAI o1 and Five Other Large Language Models: A) Accuracy and Macro-F1; B) Text-Generation Metrics

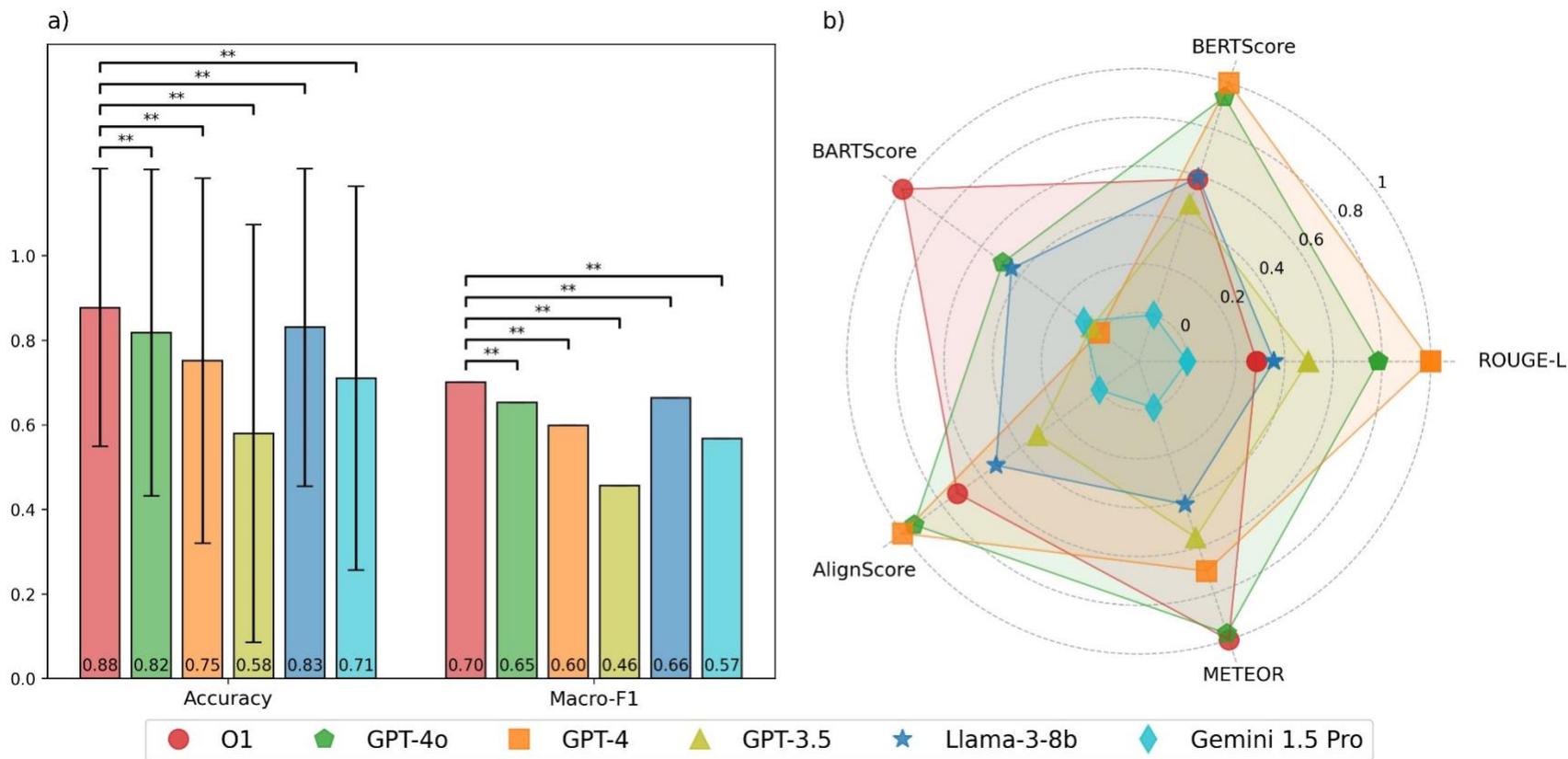

**Supplementary Tables**

Supplementary Table 1: Definitions of the Text Generation Metrics Used in this Study

| Text Generation metric | Brief explanation |
|---|---|
| ROUGE-L | Measures the overlap of the longest common sequence (LCS) between the model output and MedMCQA reference reasoning. |
| BERTScore | Measures the similarity between the tokens used in the model output and MedMCQA reference reasoning using contextual embeddings from models like BERT (Bidirectional Encoder Representations from Transformers). |
| BARTScore | Measures how well the model output aligns with the MedMCQA reference reasoning by predicting the reference reasoning from the model output and vice versa, using the BART (Bidirectional and Auto-Regressive Transformers) model. |
| AlignScore | Measures seminal and syntactic alignment between the model output and MedMCQA reference reasoning. |
| METEOR | Measures semantic alignment of the model output with the MedMCQA reference reasoning while taking into consideration stemming and synonyms-based based matching along with exact word matching. It is calculated via a combination of precision and recall. |

Supplementary Table 2: Selected Examples of MCQ Items with Sub-standard Reasonings Observed in the MedMCQA Dataset

| Example | Question | Options | Reasoning |
|---|---|---|---|
| **Reasoning consists of only the options** | A 10 yr old boy presents with b/1 chronic uveitis. Which investigation should be ordered: | Hemogram<br>X-ray of sacroiliac jt.<br>HIV test<br>HIV test (Correct) | A i.e. Hemogram; B i.e. X-ray of sacroiliac jt.; C i.e. HIV test |
| **Reasoning consists of restating the correct answer** | All are causes of white-dot syndrome except: | Fuch's heterochromic uveitis<br>VKH syndrome<br>HIV retinopathy<br>Sympathetic ophthalmitis (Correct) | Ans. Sympathetic ophthalmitis |
| **Reasoning consists string of keywords** | Characteristic visual field defect in optic chiasma lesion - | Homonymous hemianopia<br>Bitemporal hemianopia (Correct)<br>Upper quadrantanopia<br>Lower quadrantanopia | Ans, is 'b' i.e,. Bitemporal hemianopia Site of lesionVisual field defectOptic nerveBlindnessOptic cbiasmaBitemporal hemianopiaOptic tractIncongruent homonymous hemianopiaLateral geniculate bodyHomonymous hemianopiaOptic radiation (Total)Homonymous hemianopia, sometimes with macular sparingOptic radiation lower fibres (temporal lobe)Homonymous upper quadrantanopiaOptic radiation upper fibres (parietal lobe)Homonymous lower quadrantanopiaVisual cortex (anterior occipital cortex)Homonymous hemianopia with macular sparingOccipital cortex tipHomonymous macular defect |

Supplementary Table 3: Revised Categorization and Grouping of Ophthalmology Subtopics in the MedMCQA Dataset

| Topic Groups | Number | MedMCQA topic labels (N) |
|---|---|---|
| Corneal and external diseases | 534 | Abnormalities of corneal transparency (2) |
| | | Conjuctiva (32) |
| | | Conjunctiva (137) |
| | | Cornea (99) |
| | | Cornea and sclera (110) |
| | | Corneal degeneration (1) |
| | | Corneal dystrophies (4) |
| | | Diseases of cornea (30) |
| | | Disorders of conjunctiva (23) |
| | | Ecstatic conditions of cornea (4) |
| | | Immune-medicated scleritis (2) |
| | | Inflammations of Conjunctiva - Allergic and Granulomatous (16) |
| | | Inflammations of Conjunctiva – Infective (24) |
| | | Inflammations of the Cornea (29) |
| | | Keratoplasty (1) |
| | | Miscellaneous (Sclera) (9) |
| | | Sclera (11) |
| Lens | 336 | Acquired cataract (34) |
| | | Age-related cataract (3) |
| | | Cataract (28) |
| | | Cataract Surgery (6) |
| | | Congenital cataract (1) |
| | | Displacement of the lens (4) |
| | | Ectopia lentis (6) |
| | | Lens (220) |
| | | Lenses (6) |
| | | Miscellaneous (Lens) (14) |
| | | Surgical Techniques for Cataract Extraction (14) |
| Vitreous and retina | 319 | Diagnosis and Treatment of Retina (19) |
| | | Macular Disorders (1) |
| | | Retina (146) |
| | | Retina detachment (13) |
| | | Vitreous (4) |
| | | Vitreous and retina (111) |
| | | Vitreous Haemorrhage Vitrectomy (5) |
| | | Vitreous Humour, Uveal Disorders (19) |
| | | Vitreous Opacities (1) |
| Oculoplastic and orbital diseases | 256 | Anomalies in the Position of Lashes and Lid Margin (6) |
| | | Cysts and Tumors (1) |
| | | Diseases of orbit, Lids and lacrimal apparatus (146) |
| | | Graves' Ophthalmopathy, Tumors, Fractures, and Surgery (3) |
| | | Lacrimal Apparatus and Eyelid Disorders (15) |
| | | Lacrimal Gland (11) |
| | | Lid (11) |
| | | Nasolacrimal Disorders (15) |
| | | Oedema of Lids (3) |
| | | Orbit (26) |
| | | Orbit and Adnexa (10) |
| | | Orbital Disorders (7) |
| | | Orbital Inflammations (2) |
| Glaucoma | 213 | Congenital Glaucomas (5) |
| | | Glaucoma (173) |
| | | Primary Angle-Closure Glaucoma (12) |
| | | Primary Open-Angle Glaucoma and Related Conditions (14) |
| | | Secondary Glaucomas (4) |
| | | Surgical Procedures For Glaucoma (5) |

Supplementary Table 4: Subgroup Analysis of Openai O1 and Five Other Models' Performance on Items Correctly Answered by All Models (n=2,994)

| Metrics | Models | | | | | |
|---|---|---|---|---|---|---|
| | **OpenAI O1** | **GPT-4o** | **GPT-4** | **GPT-3.5** | **Llama-3-8b** | **Gemini 1.5 Pro** |
| **ROUGE-L** | 0.13±0.07 | 0.14±0.07** | **0.15±0.08**** | 0.14±0.07** | 0.13±0.07 | 0.12±0.06** |
| **BERTScore** | 0.66±0.06 | 0.68±0.05** | **0.68±0.05**** | 0.66±0.05* | 0.66±0.05 | 0.64±0.05** |
| **BARTScore** | **-4.69±1.02** | -4.78±1.04** | -4.85±1.04** | -4.77±1.01* | -4.79±1.02** | -4.84±1.01** |
| **AlignScore** | 0.18±0.13 | 0.20±0.14** | **0.21±0.16**** | 0.16±0.13** | 0.16±0.12** | 0.11±0.10** |
| **METEOR** | **0.22±0.09** | 0.22±0.09* | 0.21±0.10** | 0.22±0.09* | 0.21±0.09** | 0.20±0.08** |

*Denotes p-value<0.05 (after Bonferroni correction) and **denotes p-value<0.001 (after Bonferroni correction) for comparisons against OpenAI o1's performance

Supplementary Table 5: Subgroup Analysis of OpenAI o1 and Five Other Models' Performance on Items with Ground Truth Explanations Available for Both Correct and Incorrect Options (n=100)

| Metrics | Models | | | | | |
| --- | --- | --- | --- | --- | --- | --- |
| | OpenAI O1 | GPT-4o | GPT-4 | GPT-3.5 | Llama-3-8b | Gemini 1.5 Pro |
| **ROUGE-L** | 0.17±0.05 | **0.20±0.06**\*\* | 0.19±0.07\*\* | 0.19±0.06\*\* | 0.18±0.06\* | 0.16±0.06\*\* |
| **BERTScore** | 0.69±0.03 | **0.70±0.03**\*\* | 0.70±0.04\*\* | 0.69±0.03 | 0.69±0.03 | 0.68±0.03\*\* |
| **BARTScore** | **-4.45±0.75** | -4.54±0.76\*\* | -4.63±0.78\*\* | -4.61±0.77\*\* | -4.58±0.77\*\* | -4.54±0.85\*\* |
| **AlignScore** | 0.24±0.12 | 0.27±0.15\*\* | **0.28±0.16**\*\* | 0.18±0.12\*\* | 0.22±0.13\*\* | 0.15±0.09\*\* |
| **METEOR** | **0.27±0.07** | 0.26±0.09 | 0.24±0.10\*\* | 0.26±0.09\*\* | 0.26±0.07\*\* | 0.26±0.07\*\* |

\*Denotes p-value<0.05 (after Bonferroni correction) and \*\*denotes p-value<0.001 (after Bonferroni correction) for comparisons against OpenAI o1's performance

Supplementary Table 6: Accuracy Comparison of OpenAI o1 and Five Other Models Across the Five Major Ophthalmology Subtopics

| Sub-topic | Number of questions | Models | | | | | |
| --- | --- | --- | --- | --- | --- | --- | --- |
| | | OpenAI o1 | GPT4o | GPT4 | GPT-3.5 | Llama-3-8b | Gemini 1.5 Pro |
| **Corneal and external diseases** | 534 | **0.90±0.30** | 0.85±0.36* | 0.78±0.42* | 0.63±0.48** | 0.85±0.36* | 0.74±0.44** |
| **Lens** | 336 | **0.90±0.30** | 0.84±0.37 | 0.77±0.42** | 0.53±0.50** | 0.89±0.31 | 0.72±0.45** |
| **Vitreous and retina** | 319 | **0.88±0.33** | 0.83±0.37 | 0.76±0.42** | 0.62±0.49** | 0.85±0.36 | 0.75±0.43** |
| **Oculoplastic and orbital diseases** | 256 | **0.87±0.34** | 0.85±0.36 | 0.78±0.42* | 0.60±0.49** | 0.83±0.37 | 0.76±0.43* |
| **Glaucoma** | 213 | **0.89±0.32** | 0.84±0.37 | 0.76±0.43* | 0.58±0.50** | **0.89±0.31** | 0.69±0.46** |

*Denotes p-value<0.05 (after Bonferroni correction) and **denotes p-value<0.001 (after Bonferroni correction) for comparisons against OpenAI o1's performance

Supplementary Table 7: Comparative Performance of OpenAI o1 and Five Other Large Language Models on Text Generation Metrics Across the Five Major Ophthalmology Subtopics

| Sub-topic | Models | | | | | |
|---|---|---|---|---|---|---|
| | OpenAI o1 | GPT4o | GPT4 | GPT-3.5 | Llama-3-8b | Gemini 1.5 Pro |
| **Corneal and external diseases (n=534)** | | | | | | |
| ROUGE-L | 0.14±0.05 | **0.16±0.05**** | **0.16±0.06**** | 0.14±0.06 | 0.14±0.05 | 0.13±0.05* |
| BERTScore | **0.68±0.04** | **0.68±0.04** | **0.68±0.04** | 0.67±0.04 | 0.67±0.04 | 0.66±0.04 |
| BARTScore | **-4.90±0.77** | -5.01±0.76 | -5.08±0.75** | -5.08±0.76* | -5.05±0.73* | -5.03±0.76* |
| AlignScore | 0.19±0.12 | 0.21±0.14 | **0.22±0.15*** | 0.16±0.11* | 0.17±0.11 | 0.12±0.09** |
| METEOR | **0.23±0.07** | 0.22±0.08* | 0.20±0.08** | 0.21±0.07** | 0.21±0.07** | 0.22±0.07* |
| **Weighted Normalised Scoresⱡ** | 0.75 | **0.76** | 0.60 | 0.36 | 0.40 | 0.17 |
| **Lens (n=336)** | | | | | | |
| ROUGE-L | 0.14±0.06 | 0.15±0.06* | **0.16±0.06**** | 0.15±0.06 | 0.14±0.06 | 0.12±0.05* |
| BERTScore | 0.67±0.04 | **0.68±0.04** | **0.68±0.04** | 0.67±0.04 | 0.67±0.04 | 0.65±0.04 |
| BARTScore | **-4.92±0.92** | -5.02±0.93 | -5.08±0.94* | -5.09±0.89* | -5.07±0.91 | -5.05±0.87 |
| AlignScore | 0.19±0.12 | 0.22±0.15 | **0.23±0.15*** | 0.15±0.11** | 0.17±0.11 | 0.12±0.10** |
| METEOR | **0.23±0.08** | 0.22±0.08 | 0.21±0.09** | 0.22±0.08 | 0.21±0.08* | 0.21±0.07* |
| **Weighted Normalised Scoresⱡ** | **0.75** | 0.65 | 0.61 | 0.34 | 0.31 | 0.05 |
| **Vitreous and retina (n=319)** | | | | | | |
| ROUGE-L | 0.15±0.05 | 0.17±0.06* | **0.18±0.07**** | 0.16±0.06 | 0.15±0.05 | 0.14±0.05* |
| BERTScore | 0.68±0.04 | **0.69±0.04** | **0.69±0.04** | 0.68±0.04 | 0.68±0.04 | 0.66±0.04 |
| BARTScore | **-4.80±0.95** | -4.87±0.95 | -4.96±0.98* | -4.93±0.96 | -4.90±0.94 | -4.91±0.96 |
| AlignScore | 0.20±0.12 | 0.22±0.13 | **0.23±0.15*** | 0.17±0.12* | 0.18±0.11 | 0.13±0.09** |
| METEOR | **0.25±0.08** | 0.24±0.09 | 0.22±0.10** | 0.23±0.09 | 0.23±0.08 | 0.23±0.08* |
| **Weighted Normalised Scores‡** | 0.73 | **0.80** | 0.60 | 0.44 | 0.50 | 0.15 |
| **Oculoplastic and orbital diseases (n=256)** | | | | | | |
| ROUGE-L | 0.14±0.06 | 0.16±0.06* | **0.17±0.07**** | 0.15±0.06 | 0.14±0.06 | 0.13±0.06 |
| BERTScore | 0.68±0.04 | 0.68±0.04 | **0.69±0.04** | 0.67±0.04 | 0.67±0.04 | 0.66±0.04 |
| BARTScore | **-5.03±0.97** | -5.13±0.94 | -5.23±0.93* | -5.19±0.95 | -5.20±0.91 | -5.17±0.92 |
| AlignScore | 0.17±0.11 | **0.18±0.13** | **0.18±0.14** | 0.12±0.10** | 0.14±0.10 | 0.10±0.09** |
| METEOR | **0.24±0.08** | 0.23±0.09 | 0.22±0.09** | 0.23±0.08 | 0.22±0.08* | 0.22±0.07* |
| **Weighted Normalised Scores‡** | 0.74 | **0.76** | 0.59 | 0.37 | 0.31 | 0.07 |
| **Glaucoma (n=213)** | | | | | | |
| ROUGE-L | 0.14±0.05 | 0.16±0.06* | **0.17±0.06**** | 0.15±0.05 | 0.15±0.05 | 0.12±0.05* |
| BERTScore | **0.68±0.04** | **0.68±0.04** | **0.68±0.04** | 0.67±0.03 | 0.67±0.03 | 0.66±0.04 |
| BARTScore | **-4.83±0.84** | -4.98±0.81 | -5.05±0.79* | -5.02±0.83 | -4.99±0.81 | -4.99±0.79 |
| AlignScore | 0.19±0.10 | 0.21±0.13 | **0.22±0.14** | 0.15±0.11* | 0.17±0.10 | 0.12±0.09** |
| METEOR | **0.25±0.08** | 0.23±0.08* | 0.21±0.08** | 0.23±0.08* | 0.22±0.07** | 0.22±0.07* |
| **Weighted Normalised Scores‡** | **0.76** | 0.72 | 0.59 | 0.41 | 0.45 | 0.13 |

*Denotes p-value<0.05 (after Bonferroni correction) and **denotes p-value<0.001 (after Bonferroni correction) for comparisons against OpenAI o1's performance

‡Model performance in each text generation metric was normalized on a scale from 0 to 1, with 0 being assigned to the poorest score and 1 to the greatest, with the rest normalized proportionally to that scale. The Weighted normalized scores were calculated by taking the mean of all the scores across the five text generation metrics.

Supplementary Table 8a: Selected examples where both OpenAI o1 and GPT-4o responded correctly to the same question

| Question |
|---|
| A 52-year-old man presents to the eye clinic with painless vision loss of his right eye. He describes the visual loss as a gradual progression from blurry to total blackout over the past two hours. He has no history of prior visual problems. Past medical history is significant for a myocardial infarction three years ago. The patient takes 70mg of aspirin daily. Vital signs are normal. Physical examination reveals 20/20 vision of the left eye but no vision in the right eye. Extraocular muscles are intact. The neurologic examination is normal. The cardiac examination reveals an S4 hea sound. At the molecular level, which of the following components is essential for the first step of the visual cascade? |

| Options | | | |
|---|---|---|---|
| **A: 11-cis-retinal (correct)** | B: All-cis-retinal | C: All-trans-retinal | D: Meta-rhodopsin II |

| OpenAI o1 Output | GPT-4o Output |
|---|---|
| A | A |
| The first step of the visual cascade involves the absorption of a photon of light by rhodopsin in the photoreceptor cells of the retina. Rhodopsin is a photopigment composed of the protein opsin bound to 11-cis-retinal, which is a light-sensitive chromophore derived from vitamin A. When a photon hits rhodopsin, the 11-cis-retinal undergoes isomerization to form all-trans-retinal. This conformational change in retinal leads to a change in the shape of rhodopsin, activating it to form meta-rhodopsin II. This activated rhodopsin then initiates the phototransduction cascade by activating the G-protein transducin, eventually leading to hyperpolarization of the photoreceptor cell and transmission of visual signals to the brain. Therefore, 11-cis-retinal is essential for the first step of the visual cascade because it directly absorbs the photon and | The first step of the visual cascade involves the absorption of light by the photoreceptor cells in the retina. This process begins with the molecule 11-cis-retinal, which is bound to the protein opsin to form rhodopsin in rod cells. When 11-cis-retinal absorbs a photon of light, it undergoes a conformational change to all-trans-retinal. This change triggers a series of biochemical events that ultimately result in the generation of a nerve impulse that is sent to the brain. Therefore, 11-cis-retinal is essential for the initiation of the visual cascade. |

| undergoes the initial molecular change that triggers downstream signaling. | |

Supplementary Table 8b: Selected examples where OpenAI o1 was incorrect while GPT-4o was correct when responding to the same question

| Question | | | |
|---|---|---|---|
| Phenol red thread test is used for dry eye. This test | | | |
| **Options** | | | |
| A: Requires topical anesthetic agent | **B: Volume of tears isn't measured as it changes color on contact with tears (correct)** | C: If the color changes to blue, it depicts surface mucin deficiency | D: Requires pH meter for reading result |
| **OpenAI o1 Output** | | **GPT-4o Output** | |
| A | | B | |
| The phenol red thread test (PRTT) is a diagnostic tool used to measure tear production in patients with suspected dry eye disease. In this test, a small thread treated with phenol red dye is placed into the lower conjunctival fornix of the eye for a standardized time (usually 15 seconds). Phenol red is a pH-sensitive dye that changes color from yellow to red in the presence of the slightly alkaline tears.<br>To obtain an accurate measurement of basal (unstimulated) tear secretion, it is important to minimize reflex tearing that can occur due to ocular irritation from the thread. Therefore, a topical anesthetic agent is applied to the eye before performing the PRTT. The anesthetic numbs the ocular surface, reducing discomfort and preventing reflex tearing, which could artificially elevate tear production measurements.<br>Hence, option A is correct because the phenol red thread test requires the use of a topical anesthetic agent to accurately assess basal tear secretion without the interference of reflex tearing. | | The Phenol red thread test is a diagnostic test used to measure tear production in patients suspected of having dry eye syndrome. The test involves placing a small thread treated with phenol red dye in the lower fornix of the eye. The thread changes color from yellow to red upon contact with tears, and the length of the color change is measured to assess tear production. Unlike the Schirmer test, the Phenol red thread test does not require a topical anesthetic agent (ruling out option A), and it does not measure the volume of tears directly but rather the length of the color change (supporting option B). The test does not involve a color change to blue or indicate surface mucin deficiency (ruling out option C), nor does it require a pH meter for reading the result (ruling out option D). Therefore, option B is the correct answer. | |

Options B, C, and D are incorrect for the following reasons:

Option B: The volume of tears is indeed measured in the PRTT. The length of the thread that changes color from yellow to red correlates with the amount of tear production. The test relies on the color change upon contact with tears to assess tear volume.

Option C: In the PRTT, the phenol red dye changes color from yellow to red, not blue. A blue color change is not associated with this test, and surface mucin deficiency is not indicated by a blue color in this context.

Option D: A pH meter is not required for the PRTT. The test uses the phenol red dye's property of changing color in response to the pH of tears, but the color change is directly observed visually on the thread without the need for any instruments like a pH meter.

Supplementary Table 8c: Selected examples of responses where OpenAI o1 was correct while GPT-4o was incorrect when responding to the same question

| Question | | | |
|---|---|---|---|
| A patient is having right lateral rectus palsy. His head will be turning | | | |
| Options | | | |
| A: Towards right (correct) | B: Towards left | C: Upwards | D: Downwards |
| OpenAI o1 Output | | GPT-4o Output | |
| A | | B | |
| In the case of a right lateral rectus palsy, the right lateral rectus muscle—which is responsible for abducting the right eye (moving it outward)—is weakened or paralyzed. This weakness leads to an inability of the right eye to move fully outward, causing it to drift inward (esotropia) due to the unopposed action of the medial rectus muscle.  To compensate for the double vision (diplopia) that arises from the misalignment of the eyes, patients often adopt a compensatory head posture to realign their visual axes and minimize symptoms. Specifically, the patient will turn their head towards the side of the affected muscle—in this case, towards the right.  By turning the head to the right, the patient effectively moves their field of gaze to the left. This compensatory movement reduces the need for the weakened right lateral rectus muscle to abduct the eye, as the eyes are now positioned such that they require less outward movement to fixate on objects straight ahead. This maneuver helps maintain single binocular vision and reduces the strain on the affected muscle.  Therefore, the patient's head will be turned towards the right side to compensate for the right lateral rectus palsy and alleviate symptoms of diplopia. | | In the case of right lateral rectus palsy, the affected muscle is responsible for abducting the eye (moving it outward). When the lateral rectus muscle is paralyzed, the eye tends to drift inward (esotropia) because the opposing medial rectus muscle is unopposed. To compensate for the double vision (diplopia) that results from this misalignment, the patient will often turn their head towards the side opposite the affected muscle. This head turn helps align the eyes better and reduces the double vision. Therefore, a patient with right lateral rectus palsy will turn their head towards the left. | |

Supplementary Table 8d: Selected examples of responses where both OpenAI o1 and GPT-4o were incorrect when responding to the same question

| Question | | | |
|---|---|---|---|
| First order neuron of visual pathway | | | |
| Options | | | |
| A: Photoreceptor (correct) | B: bipolar neuron | C: lateral geniculate body | D: all of the above |
| OpenAI o1 Output | | GPT-4o Output | |
| B | | B | |
| In the visual pathway, the first order neurons are the bipolar neurons located in the retina. Photoreceptors (rods and cones) are specialized sensory receptor cells that detect light and convert it into electrical signals through a process called phototransduction. However, they are not considered neurons in the context of the neural pathway. The bipolar neurons receive input from these photoreceptors and act as the first true neurons in the pathway by transmitting signals to the retinal ganglion cells, which are the second order neurons. The axons of the ganglion cells then form the optic nerve, carrying visual information to the lateral geniculate body (LGN) in the thalamus, where the third order neurons are located. The LGN neurons then project to the visual cortex in the occipital lobe. Therefore, option B, 'bipolar neuron,' is the correct answer as it represents the first order neuron of the visual pathway. | | The first-order neurons in the visual pathway are the bipolar cells. These cells receive input from the photoreceptors (rods and cones) and transmit the visual information to the ganglion cells, which are the second-order neurons. The lateral geniculate body is part of the thalamus and is involved later in the visual pathway, specifically in the relay of information from the retina to the visual cortex. Therefore, the correct answer is B: bipolar neuron. | |